\documentclass{article}

\usepackage{PRIMEarxiv}

\usepackage[utf8]{inputenc} 
\usepackage[T1]{fontenc}    
\usepackage{hyperref}       
\usepackage{url}            
\usepackage{booktabs}       
\usepackage{amsfonts}       
\usepackage{nicefrac}       
\usepackage{microtype}      
\usepackage{lipsum}
\usepackage{fancyhdr}       
\usepackage{graphicx}       
\graphicspath{{figures/}}     
\usepackage{dcolumn}
\usepackage{amsmath}
\usepackage{booktabs}
\usepackage{caption}
\usepackage{multirow}
\usepackage{float}
\usepackage[ruled,vlined]{algorithm2e}
\DeclareMathOperator*{\argmin}{arg\,min}
\SetKwInput{KwInput}{Input}                
\SetKwInput{KwOutput}{Output}              
\newcolumntype{d}[1]{D{.}{.}{#1}}

		\title{Characterization of Frequent Online Shoppers using Statistical Learning with Sparsity}

\author{
  Rajiv Sambasivan, J{\"o}rg Schad, Arthur Keen, Christopher Woodward, Alexander Geenen, Sachin Sharma \\
  ArangoDB Inc \\
  San Mateo, CA\\
  \texttt{\{rajiv, joerg, arthur, christopher, alex, sachin\}@arangodb.com} \\
   \And
  Mark Burgess\\
  Chitek-i\\
  \texttt{mark-extern@arangodb.com} \\
}

\begin{document}
\maketitle

\begin{abstract}
		Developing shopping experiences that delight the customer requires businesses to understand customer taste. This work reports a method to learn the shopping preferences of frequent shoppers to an online gift store by combining ideas from retail analytics and statistical learning with sparsity. Shopping activity is represented as a bipartite graph. This graph is refined by applying sparsity-based statistical learning methods. These methods are interpretable and reveal insights about customers' preferences as well as products driving revenue to the store. 
\end{abstract}


	\section{Introduction}
	Understanding customers' preferences is important in the retail industry \cite{hbr_cust_pref}. This understanding is used to personalize the shopping experience for the customer. A recent survey \cite{retail_wire_survey} reports that a majority (eighty percent) of the customers are dissatisfied with the personalization of the shopping experience. This work illustrates that ideas from retail analytics and sparsity-based statistical learning methods \cite{hastie2019statistical} can be applied to learn these preferences. Statistical learning methods can be applied to explanatory, descriptive, or predictive tasks \cite{shmueli2010explain}. The perspective in this work is explanation and description as opposed to prediction. Insights about customer preferences are inferred by applying interpretable models. A bipartite graph is commonly used to represent customer shopping behavior, with customers and store inventory representing the bipartite sets. The incidence matrix of this graph captures the purchasing activities of the customers shopping at the store. Matrix factorization applied to this graph yields representations of the vertex sets of the bipartite graph in Euclidean space. These representations are called \emph{embeddings}. These representations permit us to leverage a large body of machine learning and data analysis techniques that are available for data with a vector space representation to data that has representation as a graph. Matrix factorization can be viewed through the lens of dictionary learning. Dictionary learning determines a sparse \emph{basis} for a matrix \cite{moitra2018algorithmic} \footnote{A variation, called the \emph{over-complete} dictionary, where the opposite is true also exists. However, in this application, a sparser basis is determined.}. In the context of this application, this sparse basis yields insights about purchase behavior and customer preferences and is the characterization of frequent shoppers. This characterization can be represented as a graph and stored in a \emph{graph database}. The embeddings and the graph characterization can be used to support ad-hoc analytic queries about frequent shoppers as well as developing applications such as recommendation systems. The data for this work comes from a real-world online store. Significant preprocessing and analysis are required to develop the representation that yields the characterization and the embeddings. The description of the data for this work is provided in section \ref{sec:data_characterization_and_problem_desc}. The methodology applied to develop the characterization is discussed in section \ref{sec:methodology}.  Research related to this work is discussed in section \ref{sec:related_work}. The relevant theoretical background is discussed in section \ref{sec:theoretical_background}. The results from applying the methodology are discussed in section \ref{sec:results}. The conclusions from this study are discussed in \ref{sec:conclusion}. 	
	
	\section{Data Characterization and Problem Description}\label{sec:data_characterization_and_problem_desc}
	The data for this work comes from the UCI machine learning repository \cite{dua_2019} and represents the line items of invoices of sale from an online store in the United Kingdom selling novelty gift items. A line-item captures the sale details associated with a single inventory item and the identity of a customer if the customer has registered with the store. Customers need not register to purchase at the store. The store serves both retail and wholesale customers. The data reflects a sample of transactions between December 2010 and December 2011 and reveals $4337$ registered customers. Of these, $447$ customers have made five or more retail purchases at the store. Most \emph{retail} customers had made less than four purchases at the store. Retail customers who make five or more purchases at the store are determined to be \emph{frequent shoppers}. This group has a shopping history at the store. The characterization of this group is developed in this work.  The characterization of the other two groups, \emph{wholesale} customers, and, customers who have made less than four purchases at the store, the \emph{infrequent shoppers}, is not developed in this work. 
	
	\begin{figure}[h] \label{fig:freq_cust_bipartite_graph}
		\centering
		\includegraphics[scale=0.35]{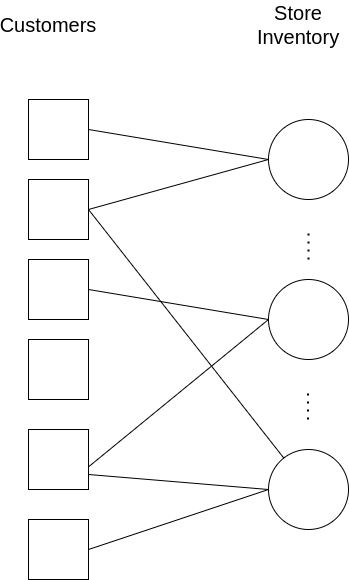}
		\caption{Bipartite graph representation of frequent shoppers.}
		\label{fig:freq_cust_bipartite_graph}	
	\end{figure}
	The purchase activity of the frequent shopper group can be represented as a bipartite graph as shown in Figure \ref{fig:freq_cust_bipartite_graph}. The \emph{incidence matrix} of this graph captures the purchasing activity of the frequent shoppers. The edges of this graph are \emph{bi-directional}.\footnote{The edges from a customer represent his or her shopping activity at the store. Tracing the edges from an inventory item provides us the customers purchasing it.} \emph{The incidence matrix captures the relationship between a customer and the store inventory. Each row of the matrix captures the purchases of a frequent shopper. This matrix is similar to the document-term incidence matrix encountered in information retrieval \cite{nlp_ir_online_ref}\cite{DBLP:books/daglib/0021593}}. The columns of the incidence matrix are the store inventory items purchased by the frequent shopper group. If a customer has purchased an item, then the corresponding entry in the matrix contains the amount spent by the customer on the item. If a customer has not purchased an item, that entry has a $0$ value. The frequent shopper group has purchased $2664$ items. Since there are $447$ frequent shoppers, the incidence matrix has dimensions of $(447 \times 2664)$. The entries of the incidence matrix are \emph{non-negative}. In this work, matrix factorization is applied to develop a concise description of the purchase activity of frequent shoppers. This requires us to determine a suitable \emph{basis} for this matrix. A preprocessing step, feature selection, can enhance the conciseness of the results.  
	
	The notion of \emph{Recency-Frequency-Monetary Value}(RFM) is used in retail analytics to capture the value of a customer \cite{dua_2019}. Recency, Frequency, and Monetary Value are attributes associated with a shopper. The \emph{Recency} attribute captures the recency of a shopper's visit to the store. The \emph{Frequency} attribute captures the number of times the shopper has made purchases at the store. The \emph{Monetary Value} captures the total amount of money spent by the shopper at the store. In this work, this idea is applied to perform feature selection. Thus we eliminate purchase activity that is considered irrelevant. A derived attribute, \emph{RFM score}, denoted by $\gamma_i$, is defined for the $i^{th}$ customer as in Equation \ref{eqn:weighted_rfm}.
	
	\begin{equation}\label{eqn:weighted_rfm}
	\gamma_i = w_{recency} Recency_i + w_{frequency} Frequency_i + w_{monetary\ value} Monetary\ Value_i
	\end{equation}
	where:
	\begin{itemize}
		\item $w_{recency}$: weight assigned to the Recency attribute
		\item $w_{frequency}$: weight assigned to the Frequency attribute
		\item $w_{monetary\ value}$: weight assigned to the Monetary Value attribute  
	\end{itemize}

	The feature selection step yields a denser and smaller representation of a customer. \emph{The incidence matrix after the feature selection step represents the purchase activity of the frequent shoppers that account for customer value}. Matrix factorization applied to this matrix yields vector representations of the customers and store inventory items. These vector representations called \emph{embeddings} can be used for several novel applications, for example, examining the similarity of customers, searching for customers and inventory items in a database, etc. Viewing the factorization of the matrix as learning a sparse basis provides a useful perspective \cite{eggert2004sparse} \cite{lecturenotes_nicolai}. This perspective yields insights into purchase behavior and customer preferences. This is discussed in section \ref{sec:nmf_theory}. The methodology used to perform feature selection and apply matrix factorization is discussed in section \ref{sec:methodology}

	\section{Methodology}\label{sec:methodology}
\begin{figure}[H]
		\centering
		\includegraphics[width=0.9\linewidth]{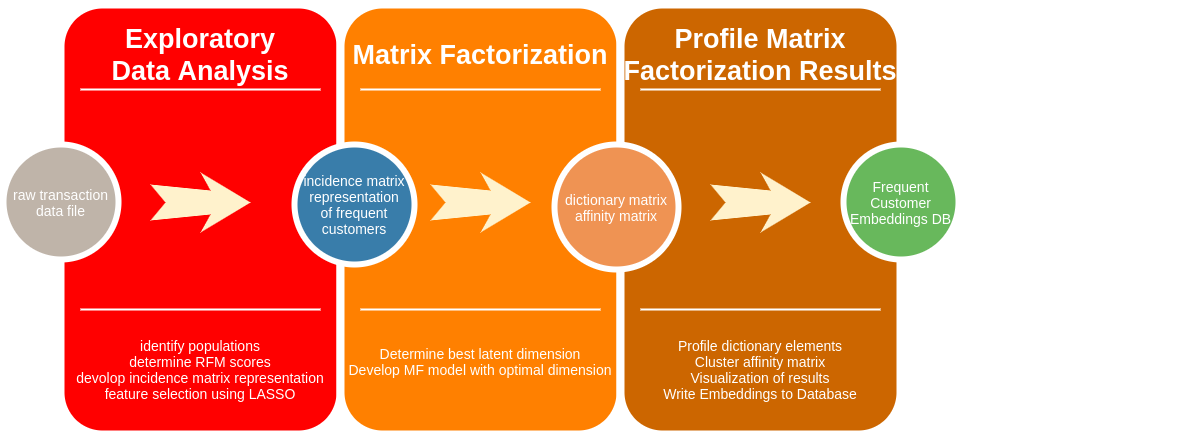}
		\caption{Methodology to develop characterization of frequent shoppers.}
		\label{fig:mf_methodology}	
	\end{figure}

	Figure \ref{fig:mf_methodology} is a schematic diagram. This process is fairly typical in data mining. The first step of the process is \emph{exploratory data analysis}. As part of this step, the characteristics of the dataset are determined. This includes determining the types of attributes, the different types of populations in this dataset, and the characteristics of each population. As discussed in section \ref{sec:data_characterization_and_problem_desc}, in this work, we are interested in the group of customers who have made five or more purchases at the store. This group is called the \emph{frequent shoppers} group. The purchasing activities of this group have a representation as a bipartite graph. The incidence matrix of this representation is determined. A \texttt{colab} \cite{google_colab} notebook  that provides a detailed implementation of the above steps is available at \cite{retail_eda_1}.
	
	The incidence matrix is high dimensional. Since a typical customer purchases, a small set of inventory items, the row capturing the customer's purchases has many zeros. The RFM score discussed  in section \ref{sec:data_characterization_and_problem_desc} is used to select the features that capture the customer's value. A linear model that regresses the customers' value (RFM score) versus their purchases is used. In particular, a LASSO model is used for feature selection. The raw RFM scores are not distributed normally. Preprocessing the response to be normally distributed is a common preprocessing step in the development of linear models \cite{fox2019regression} (see section \ref{sec:rfm_transformation}). A Box-Cox transformation \cite{box1964analysis} is used for this purpose. The details of developing the LASSO model are provided in section \ref{sec:lasso}. The LASSO model provides us a smaller and denser customer representation. This representation uses a much smaller set of inventory items to capture the shopping activity of the frequent shoppers. This representation determines a bipartite graph that captures most of the customer value of the frequent shopper segment. The second step of the process applies matrix factorization to the incidence matrix of this graph. Matrix factorization determines a Euclidean representation of the nodes of the bipartite graph representing customer purchase activity (after feature selection). This is achieved by factoring the matrix into a \emph{dictionary} matrix and an \emph{affinity} matrix. The details of the factorization and the interpretation of these matrices are provided in section \ref{sec:nmf_theory}. This section discusses a method to select the hyper-parameter associated with the Euclidean representation. Analysis of the results from matrix factorization is discussed in section \ref{sec:results}. The relevant theoretical background is discussed in section \ref{sec:theoretical_background}.
	
	\section{Related Work}\label{sec:related_work}
	The data for this work has been analyzed by \cite{chen2012data} by treating the \emph{Recency}, \emph{Frequency} and \emph{Monetary-value} as the primary attributes that represent the customer. In contrast, this work represents a customer by his or her purchases. As results from this work will show, such a representation yields insights about customer taste and store inventory. In \cite{chen2012data}, the representation of the customer is clustered by K-Means \cite{lloyd1982least}. Specific segments that are deemed as valuable are then profiled using a decision tree \cite{breiman2017classification}. This work also takes a rigorous data analysis perspective before customer characterization \cite{retail_eda_1}. Distinct populations of customers are defined. The scope of this work is limited to one of the identified populations, \emph{frequent shoppers}. The analysis methodology is based on refining the characterization in stages using well-established ideas in statistical learning. In \cite{chen2012data} clustering is used to identify groups with similar \emph{customer value}. In this work, clustering is used to identify groups with similar \emph{shopping preferences}. This work uses a density-based approach to clustering as opposed a distance-based clustering.\\
	Real-world dataset present with noise. This refers to points that do not exhibit clustering tendency from a standpoint of distance. They are not close to a neighbor. However, because we pick a finite number of clusters, such points are assigned a cluster. As noted in \cite{mcinnes2017accelerated}, density-based approaches can identify datasets with noise. Results observed with this data indicate that noise was indeed present. In this work, feature selection based on customer value is performed before the matrix factorization step. Exploring supervised matrix factorization \cite{austin2018fully} \cite{choo2015weakly} that can use the notion of customer value in selecting a sparse representation of a customer is an area of future work. Comparison of customer preferences identified in this work with other latent-factor models, such as Latent-Dirichlet Allocation \cite{blei2003latent}, is an area of future work. Exploring graph reduction techniques, such as graph coarsening \cite{andreas_graph_coarsening}, to characterize the frequent shopper group is also an area of future work.

	\section{Theoretical Background}\label{sec:theoretical_background}
	The theoretical background about the ideas discussed in section \ref{sec:methodology} are provided in this section. In the analysis, let $n$ be the number of customers in the frequent shopper group and $m$ be the number of items in the store inventory before feature selection. Results obtained using the ideas discussed in this section are reported in section \ref{sec:results}. The LASSO and non-negative matrix factorization discussed in this section determine a \emph{sparse} basis. The LASSO determines a \emph{sparse} basis that is associated with the solution of a linear model. Matrix factorization determines a \emph{sparse} basis that can be used to approximate a matrix. Seeking sparsity is a common theme in the solution of both these problems.
	\subsection{RFM score transformation}\label{sec:rfm_transformation}
	Transforming the RFM score to be normally distributed provides several advantages. Working with a normally distributed random variable permits us to analyze customer value distribution using the properties of a well-known distribution. Maximum likelihood properties based on the Normal distribution have been extensively studied for linear models. By transforming the RFM score to be normally distributed, we have a known framework for the evaluation of the theoretical properties of the model \cite{fox2019regression}. The Box-Cox transformation is used to transform the raw RFM score to a normally distributed value. In this work, the implementation of the Box-Cox transform in \texttt{scikit-learn} \cite{scikit-learn} is used. The Box-Cox transformation determines a constant, $\lambda$, that transforms the raw RFM score of the $i^{th}$ customer, $\gamma_{i}$, to a value $\gamma^{\prime}_{i}$ using Equation \ref{eqn:bctransform}.

	\begin{equation}\label{eqn:bctransform}
	\gamma^{\prime}_{i} = \gamma^{\lambda}_{i} = \left \{
	\begin{aligned}
	&\frac{(\gamma^{\lambda}_{i}-1)}{\lambda}, && \text{for}\ \lambda \neq 0 \\
	&\log(\gamma_{i}), && \text{for}\  \lambda = 0
	\end{aligned} \right.
	\end{equation}
	where:
	\begin{itemize}
		\item $\gamma^{\prime}_{i}$: is the Box-Cox transformation of the raw RFM score, $\gamma_{i}$, of the $i^{th}$ customer
		\item $\lambda$: is a parameter determined from the Box-Cox transformation 
	\end{itemize}
	A linear model is used to capture the relationship between the transformed RFM score of the $i^{th}$ customer and their purchases of store inventory as described by Equation \ref{eqn:linearcustmodel}. 
	\begin{equation}\label{eqn:linearcustmodel}
	\gamma^{\prime}_{i} = \beta_0 + \beta_{1} p_{1i} +  \beta_{2} p_{2i} + \ldots +  \beta_{m} p_{1m} + \epsilon_i
	\end{equation} 
	where:
	\begin{itemize}
		\item $p_{i1},\ldots p_{im}$: are the purchases of store inventory items $1 \ldots m$ by the $i^{th}$ customer
		\item $\epsilon_i$: are independent identically distributed errors from a normal distribution, $\mathbf{N}(0,\sigma^2)$. 
	\end{itemize}
	
	A linear relationship between the transformed RFM score and customer purchases, as represented in Equation \ref{eqn:linearcustmodel}, is simple and interpretable. In matrix form, this relationship can be written as
	\begin{equation}\label{eqn:matrix_cust_rfm}
	\boldsymbol{\gamma^{\prime}} = \mathbf{P}.\boldsymbol{\beta} +  \boldsymbol{\epsilon}
	\end{equation}
	where:
	\begin{description}
		\item $\boldsymbol{\gamma^{\prime}}$: is a $n \times 1$ matrix of transformed customer RFM scores
		\item  $\mathbf{P}$: is a $n \times m$ matrix of customer purchases of store inventory
		\item $\boldsymbol{\beta}$: is a $ m \times 1$ matrix of linear model coefficients
		\item $\boldsymbol{\epsilon}$: is a $ n \times 1$ matrix of errors (from $\mathbf{N}(0,\sigma^2)$).
	\end{description}
	
	However, this poses a problem. We have $447$ customers ($n=447$) and $2664$ inventory items ($m=2664$). This is an example of an \emph{underdetermined} system. We have a linear system with more unknown's ($\boldsymbol{\beta}$) than equations (number of customers). \emph{Regularization} is a technique to find solutions to an underdetermined system. The \emph{Least Absolute Shrinkage and Selection Operator} (LASSO) \cite{tibshirani1996regression} can be used for this purpose. The details of the LASSO are provided in section \ref{sec:lasso}. When the predictor variables are drawn from a continuous distribution, the LASSO produces a unique solution (see \cite{tibshirani2013lasso}).\footnote{In this case, the user's spend on the various items of the store inventory are the predictor variables. The notion that these are continuous random variables is a reasonable one.}
	\subsection{The Least Absolute Shrinkage and Selection Operator (LASSO)} \label{sec:lasso}
	The LASSO is a regression analysis method that can be used for regularization and feature selection with linear models. In this work, LASSO is used to incorporate regularization and feature selection into the linear model described by Equation \ref{eqn:matrix_cust_rfm}. The LASSO determines a sparse set of coefficients for a linear model by minimizing the objective function shown in Equation \ref{eqn:rfmpurchases}. This formulation finds a \emph{basis} that permits us to explain the variation in customer value with a small number of inventory elements. This approach is also called \emph{basis pursuit} \cite{chen1998application} in the literature. A detailed exposition of the LASSO that provides an excellent account of its formulation, solution and properties is available in \cite{hastie2019statistical}.
	\begin{equation}\label{eqn:rfmpurchases}
	\min_{\beta \in \mathcal{R}^{m}} \quad \|\boldsymbol{\gamma^{\prime}} - \mathbf{P}.\boldsymbol{\beta} \| + \alpha \|\boldsymbol{\beta}\| 
	\end{equation}
	where:
	\begin{itemize}
		\item $\boldsymbol{\gamma^{\prime}}$,  $\mathbf{P}$ and  $\boldsymbol{\beta}$: Are defined as in Equation \ref{eqn:matrix_cust_rfm}
		\item $\alpha$: is a regularization hyper-parameter. 
	\end{itemize}
	
\begin{figure}[ht]
		\centering
		\includegraphics[scale=0.33]{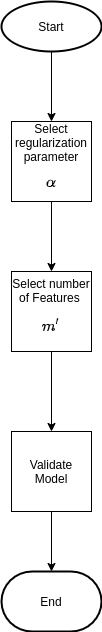}
		\caption{Using the LASSO for feature selection.}
		\label{fig:lasso_flowchart}	
	\end{figure}

	A schematic illustration of the process used to perform feature selection using the LASSO model is shown in Figure \ref{fig:lasso_flowchart}. The details of each step of the process illustrated in Figure \ref{fig:lasso_flowchart} are as follows.
	
	\begin{enumerate}
		\item Select Regularization Parameter, $\alpha$: Applying the LASSO model for feature selection requires the selection of the regularization parameter, $\alpha$. This is selected by considering a range of values of $\alpha$ and picking the value that produces the lowest cross-validation error \cite{kfoldcrossval}. 
		\item Select the number of features: Once the $\alpha$ parameter has been selected, the LASSO model can be computed and used to perform feature selection. The LASSO solution drives the coefficients in the linear model, $\beta$, to zero. The absolute value of the coefficient is a direct measure of its ability to explain customer value. An ordering of these coefficients in descending order provides a measure of feature importance of an item of inventory in explaining the customer value of the frequent shopper group. Many of these coefficients are very small and do not explain customer value. We only want those coefficients that can account for explaining a meaningful amount of customer value. To determine the number of coefficients to select from the model, the following experimental procedure is used. A portion of the dataset is held out as a test set. The solution from the lasso is initially selected as the set of features to include in the model. A linear model is developed with these features and its performance on the test set is noted. The smallest coefficient is dropped and a model is developed with the trimmed feature set. The performance of the model on the test set is noted. This process of dropping the feature associated with the smallest coefficient is repeated and model performance is noted. We then plot the number of features included in the model versus test set performance. This plot reveals the optimal number of features to include in the model. The experimental procedure is described in Algorithm \ref{alg:exp1}. \emph{This procedure yields a set of $m^{\prime}$ coefficients that can explain most of the value in the frequent shopper group. Results showed that $m^{\prime} \ll m$. In other words, a small set of store inventory items account for most of the customer value}. The details are discussed in section \ref{sec:results}, the relevant fact is that such an experiment reveals that there diminishing returns to including a feature in terms of its ability to explain customer value. The above procedure gives us an experimental estimate of the number of store inventory items that contribute towards explaining customer value.
		\begin{algorithm}[H]
			\DontPrintSemicolon
			
			\KwInput{The LASSO solution, $\beta_{\text{LASSO}}$, The hold-out test set}
			\KwData{The frequent customer training data}
			plot data = [ ]
			\tcc{Compute hold-out test error with LASSO solution}
			num\_features = number of features in the LASSO solution\\
			fit\_linear\_model(num\_features)\\
			add\_holout\_error\_to\_plot\_data()\\

			\tcc{Observe the hold-out error as we drop features from the feature set for the linear model}
			
			\While{ num\_features $>0$}
			{	
				fit\_linear\_model(num\_features)\\
				add\_holout\_error\_to\_plot\_data()\\
				\tcc{Drop one feature}
				drop\_feature\_with\_smallest\_coefficient()
			}
			
			plot\_experiment\_curve(plot data)
			
			\caption{An experiment to select number of features from the LASSO solution}
			\label{alg:exp1}
		\end{algorithm}

		\item Validate that the assumption of a linear model:  We still need to assess if the proposed linear model for estimating customer value is a reasonable choice. This assessment is done by:
		\begin{enumerate}
			\item Plotting the model estimates of customer value versus the actual customer value for the test set.
			\item Examining a probability-probability plot of the errors from the test set against a normal distribution to evaluate if normality of errors is a reasonable assumption.
		\end{enumerate} 
	\end{enumerate}
	
	A \texttt{colab} \cite{google_colab} notebook  that provides a detailed implementation of feature selection using the LASSO as illustrated in Figure \ref{fig:lasso_flowchart} is available \cite{retail_eda_2}.

	\subsection{Non-Negative Matrix Factorization}\label{sec:nmf_theory}
\begin{figure}[H]
		\centering
		\includegraphics[scale=0.25]{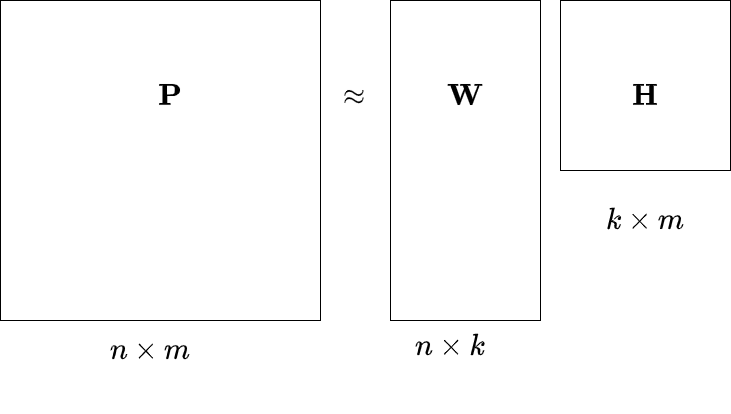}
		\caption{Matrix factorization factors, number of customers ($n$), number of inventory items ($m$), latent dimension $k$.}
		\label{fig:mf_factors}	
	\end{figure}

	Feature selection performed using the LASSO provides a small number of features, $m^{\prime}$ that can account for most of the customer value. The matrix representing customer purchases after feature selection, $\mathbf{P}^{\prime}$, has fewer columns ($m^{\prime}$) than the original matrix of purchases, $\mathbf{P}$, which has $m$ columns. \emph{The matrix $\mathbf{P}^{\prime}$ is the incidence matrix of a bipartite graph that captures the purchasing activity of the frequent shopper group}. It should be noted that the entries $p_{ij}$ of $\mathbf{P}^{\prime}$ represent the purchase of item $j$ by customer $i$. This is a strictly non-negative quantity. Non-negative matrix factorization can be used to factorize the matrix $\mathbf{P}^{\prime}$ into factors $\mathbf{W}$ and $\mathbf{H}$ such that $ \mathbf{P}^{\prime} \approx \mathbf{W}.\mathbf{H}$, as shown in Figure \ref{fig:mf_factors}. 
	
	The matrix $\mathbf{W}$ of dimension $n \times k$ is tall and narrow, the matrix $\mathbf{H}$ of dimension $k \times m$, is shallow and wide. Matrix factorization approximates the purchase of the $i^{th}$ customer as the product of two factors, a dictionary element $h_{km}$ and an affinity element $w_{ik}$, such that the purchases of the customer can be expressed as shown in Equation \ref{eqn:mf_purchase_approx}.
	\begin{equation}\label{eqn:mf_purchase_approx}
	p_i \approx w_{ik}. h_{km}
	\end{equation}
	The dimension $k$ is a hyper-parameter to the method and is determined through an imputation experiment. A third of the entries of the matrix $\mathbf{P}^{\prime}$ are held out as a test set. A range of values, from $2$ to $20$, are used as candidate values for $k$. For each value of $k$, the mean square error (MSE) in predicting the entries of the test set is computed. The value of $k$ that produces the lowest mean square error on the test set is selected as the optimal value of the latent dimension. We now present two perspectives of matrix factorization to motivate the essential aspects of the developing the solution.
	
	\begin{enumerate}
		\item A dictionary learning perspective: The matrix $\mathbf{H}$ represents a dictionary of purchases. It represents a \emph{basis} for a latent space of dimension $k$. The purchases of any customer $i$ can be expressed as a linear combination of the elements of the dictionary, as expressed by Equation \ref{eqn:mf_purchase_approx}. Consequently, matrix factorization yields a dictionary of purchases that we can group the store inventory into and a matrix of customer affinities towards these dictionary elements. The dictionary elements and the affinities have vector space representations. The rows of the matrix $\mathbf{W}$, the affinity matrix, represent a $k$ dimensional representation of the customer. The columns of the dictionary matrix $\mathbf{H}$ represent the $k$ dimensional representation of the store inventory items. These representations can be used by applications such as similarity search\cite{johnson2019billion}. By examining the rows of the matrix $\mathbf{H}$ the inventory items that dominate each of the $k$ dictionary items can be determined. An entry $h_{ij}$ of a dictionary element can be viewed as the contribution of the $j^{th}$ inventory item to the $i^{th}$ dictionary element. A high value implies that the $j{th}$ inventory item contributes strongly towards the $i^{th}$ dictionary item. By examining the entries of $\mathbf{H}$ with a row-wise scan, we can identify the inventory items that dominate each dictionary element. Similary, an entry $w_{ij}$ can be interpretted as the affinity that customer $i$ has towards dictionary element $j$. By examining the $i^{th}$ row of  $\mathbf{W}$, we can identify the preferences of the $i^{th}$ customer towards the dictionary elements. \footnote{There elements of the dictionary are also called \emph{atoms}\cite{hamon2016convex} and \emph{parts} \cite{lee1999learning} in the literature. Similarly, the affinity component is called \emph{activations}\cite{essid2014tutorial} in the literature.}
		
		The problem of determining a  non-negative matrix factorization is \emph{ill-posed}. There can multiple factorizations for a given data matrix $\mathbf{P}^{\prime}$. There have been several approaches to mitigate the non-uniqueness of the factorization by imposing different kinds of constraints, for example, using convexity \cite{DBLP:journals/pami/DingLJ10} or sparsity \cite{fevotte2011algorithms}. The perspective in this work is explanation and description. Occam's razor \cite{DBLP:journals/datamine/Domingos99}\cite{DBLP:journals/cacm/Domingos12} is applied in this work. We seek simple or sparse explanations. The sparsity constraint is imposed in the optimization model used to determine the factorization. This is discussed next.

		\item An optimization perspective: Non-negative matrix factorization applied to the matrix $\mathbf{P}^{\prime}$ requires us to determine matrices $\mathbf{H}$ and $\mathbf{W}$ such that $ \mathbf{P}^{\prime} \approx \mathbf{W}.\mathbf{H}$. These matrices are determined by framing matrix factorization as an optimization problem. In this work the implementation of matrix factorization in \texttt{scikit-learn}\cite{scikit-learn} is used. This implementation uses a flexible cost function, called \emph{$\beta$ divergence}\cite{fevotte2011algorithms}, to compute the matrix factorization. This cost function is given by Equation \ref{eqn:nmfobjective}. Regularization using a mix of both $\ell^{1} and \ell^{2}$ penalties is possible with this implementation by supplying the algorithm with a ratio parameter, $\ell^{1}_{ratio}$. The sparsity constraint is imposed by the choice of the parameter $\alpha_{m}$. In this work, the $\ell^{1}_{ratio}$ parameter and the $\alpha_{m}$ parameter are selected by a grid search. A range of values for the $\ell^{1}_{ratio}$ and $\alpha_{m}$ parameters are considered as candidate choices. The set of values that minimized the imputation error in the experiment described above was used for developing the matrix factorization model.     
		
		\begin{equation}\label{eqn:nmfobjective}
		\begin{aligned}
		\argmin_{\mathbf{W} \in \mathcal{R}^{n \times k},\ \mathbf{H} \in \mathcal{R}^{k \times m}} \quad  & \Bigg\{ \frac{1}{2}\|\mathbf{P}^{\prime} - \mathbf{W}.\mathbf{H} \| + 
		\alpha_{m} . \ell^{1}_{ratio} .\|\mathbf{W}\|_{\ell^1} + \\ & \alpha_{m} . \ell^{1}_{ratio} .\|\mathbf{H}\|_{\ell^1} + \frac{1}{2} .\alpha_{m}.(1-\ell^{1}_{ratio}) \|\mathbf{W}\|^{2}_{Frob} + \\ & \frac{1}{2} .\alpha_{m}.(1-\ell^{1}_{ratio}) \|\mathbf{H}\|^{2}_{Frob} \Bigg \}
		\end{aligned}
		\end{equation}
		where:
		\begin{itemize}
			\item $\mathbf{P}^{\prime}$: is a matrix of purchases after feature selection, of size $n \times m^{\prime}$ 
			\item $\mathbf{H}$: is a dictionary matrix of purchases, of size $m^{\prime} \times k$
			\item $\mathbf{W}$: is a matrix of customer affinities, of size $n \times k$
			\item $\alpha_m$: is a constant that multiplies the regularization terms.
			\item $ \ell^{1}_{ratio}$: a mixing parameter that specifies the ratio of  $\ell^{1} and \ell^{2}$ penalties to use with the optimization.
			\item $\|\mathbf{W}\|_{\ell^1}, \|\mathbf{H}\|_{\ell^1}$: are element-wise $\ell^{1}$ norms, $\sum_{i,j}\|w_{ij}\|$ and $\sum_{i,j}\|h_{ij}\|$ of $\mathbf{W}$ and $\mathbf{H}$.
			\item $\|\mathbf{W}\|^{2}_{Frob} , \|\mathbf{H}\|^{2}_{Frob}$: are Frobenius norms,  $\sum_{i,j} w_{ij}^{2}$ and $\sum_{i,j} h_{ij}^{2}$ of $\mathbf{W}$ and $\mathbf{H}$.     
		\end{itemize}
		
	\end{enumerate}
	
\begin{figure}[H]
		\centering
		\includegraphics[width=0.7\linewidth]{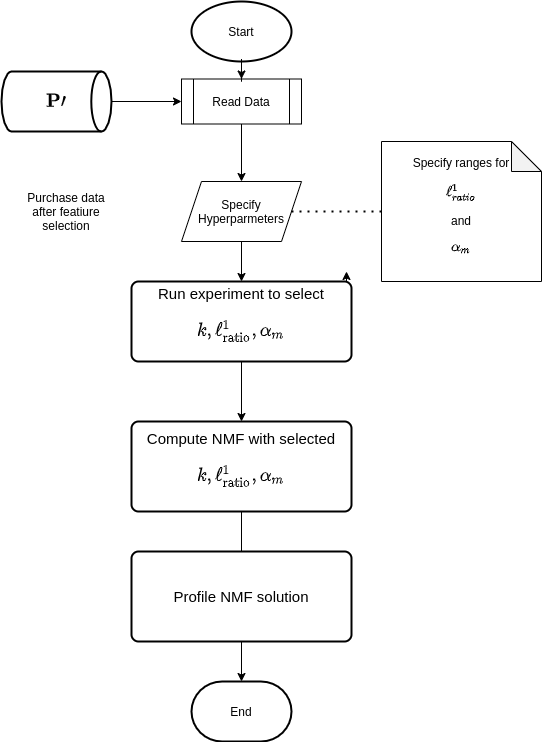}
		\caption{Developing embeddings and profiling customer preferences with matrix factorization.}
		\label{fig:mf_flowchart}	
	\end{figure}

	A flow chart illustrating the process of applying matrix factorization to develop the embeddings of the frequent shopper graph and develop a profile of customer preferences is shown in Figure \ref{fig:mf_flowchart}.
	
	\section{Results}\label{sec:results}
	This section presents the key results from applying the methodology discussed in section \ref{sec:methodology}. The \emph{exploratory data analysis} step identifies the frequent shopper group. The details of \emph{exploratory data analysis} are available at \cite{retail_eda_1}. The purchasing activities of the frequent shopper group have a representation as a bipartite graph. The incidence matrix of this graph is used to develop the characterization of the frequent shopper group. The incidence matrix developed from the raw data set provides a high-dimensional representation of the customer. A typical \emph{retail} customer purchases a small number of inventory items. Therefore, most entries associated with the customer in the incidence matrix are zeros. Feature selection using the LASSO \cite{tibshirani1996regression} is used to develop a denser representation of the customer as discussed in section \ref{sec:lasso}. The results from this step are discussed in section \ref{sec:results_lasso}. Matrix factorization provides a vector representation of the nodes of the frequent shopper graph, i.e, of customers and the store inventory. The results from the matrix factorization step are provided in section \ref{sec:results_nmf}. 
	
	\subsection{Feature Selection with the LASSO}\label{sec:results_lasso}
	A flow chart illustrating the steps involved in using the LASSO to perform feature selection is shown in Figure \ref{fig:lasso_flowchart}. A detailed implementation of this flowchart as a \texttt{colab} notebook is available at \cite{retail_eda_2}. This section captures the salient results from executing the process depicted in Figure \ref{fig:lasso_flowchart}. The theoretical background is discussed in section \ref{sec:lasso}. The parameters used for the process depicted in Figure \ref{fig:lasso_flowchart} and results from the feature selection process are summarized in Table \ref{tab:lasso_fs_param_results}.
	
	\begin{table}[ht]
		\centering
		\caption{Parameters and results from feature selection.}
		\begin{tabular}{ll}
			\toprule
			FS Parameter/Result &                        Value \\
			\midrule
			RFM weights &     $w_r=0.15,$ \\ & $w_f=0.15,$\\ &  $w_m=0.7$ \\
			Regularization parameter, $\alpha$ &                         0.22 \\
			$R^2$ (explained variance) &                        0.855 \\
			optimal number of features &                           75 \\
			training MSE &                        0.137 \\
			test MSE &                        0.326 \\
			\bottomrule
		\end{tabular}\label{tab:lasso_fs_param_results}
	\end{table}
	
\begin{figure}[ht]
		\centering
		\includegraphics[scale=0.33]{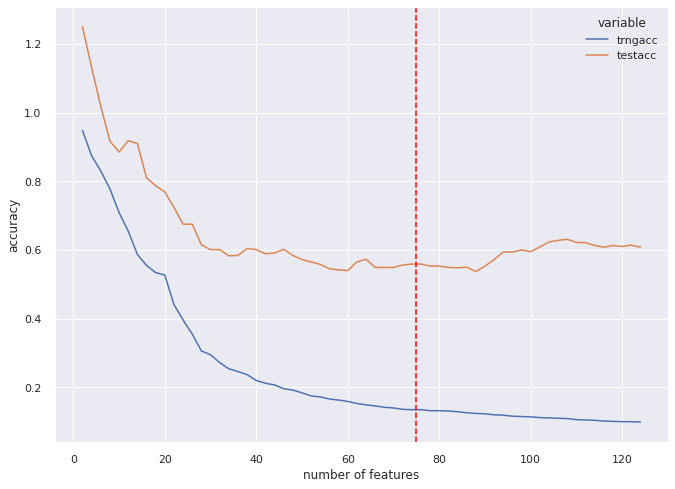}
		\caption{Optimal number of features to select from LASSO solution.}
		\label{fig:num_features_optimal}	
	\end{figure}

	The LASSO drives the coefficients of the coefficients, $\beta$, in Equation \ref{eqn:rfmpurchases} to zero. Many of these coefficients are very small and do not contribute in a meaningful manner to explaining variance in customer value ($\gamma^{\prime}$). The experiment described by Algorithm \ref{alg:exp1} is used to select those features that contribute to explaining the variation in customer value. The results from this experiment are shown in Figure \ref{fig:num_features_optimal}.
	
\begin{figure}[ht]
		\centering
		\includegraphics[scale=0.33]{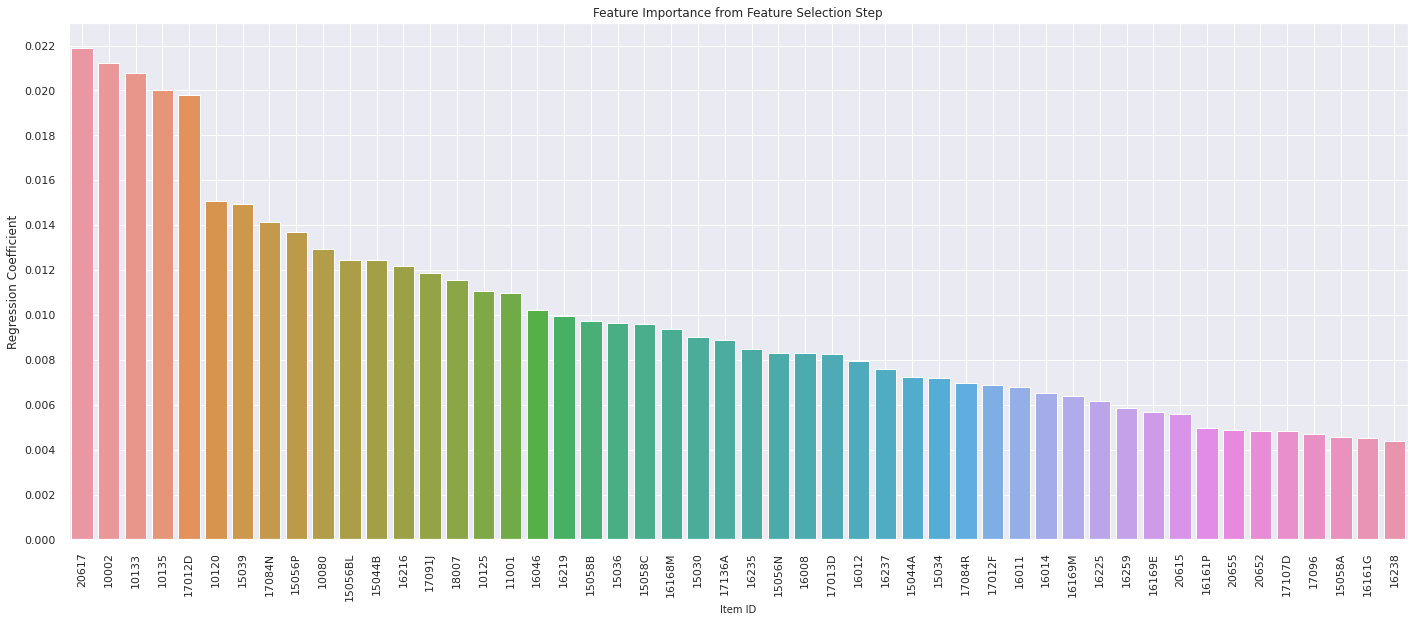}
		\caption{Feature importance of the first 50 features.}
		\label{fig:feature_importance}	
	\end{figure}

	Since a linear model is used, the magnitude of the coefficients, $\beta$, is a direct indication of an inventory item in determining customer value. An inspection of Figure \ref{fig:num_features_optimal} shows that the first $50$ features account for explaining most customer value. There are diminishing returns in including the inventory items associated with the coefficients after the $50^{th}$ coefficient in the representation of the customer. About $75$ inventory items can explain about $86\%$ of the variation in customer value. Therefore, purchases of these $75$ store items are used to characterize customer behavior in subsequent analysis. The feature importance of the first $50$ features is shown in Figure \ref{fig:feature_importance}.
	
A description of the top ten items in terms of feature importance are shown in Table \ref{tab:t10_feat_importance_desc}. \emph{It should be noted that the features in Table \ref{tab:t10_feat_importance_desc} are features that are determined important based on value (RFM score). These are the features that account for customer value in contrast to customer taste. The dictionary learned from matrix factorization, discussed in the next section, provides insights about customer taste.}

	\begin{table}[ht]
		\centering
		\caption{Description of the top 10 items in feature importance.}
		\begin{tabular}{ll}
			\toprule
			ITEM ID &                       Description \\
			\midrule
			20617 &       FIRST CLASS PASSPORT COVER  \\
			10002 &       INFLATABLE POLITICAL GLOBE  \\
			10133 &      COLOURING PENCILS BROWN TUBE \\
			10135 &      COLOURING PENCILS BROWN TUBE \\
			17012D &  ORIGAMI ROSE INCENSE/CANDLE SET  \\
			10120 &                      DOGGY RUBBER \\
			15039 &                    SANDALWOOD FAN \\
			17084N &             FAIRY DREAMS INCENSE  \\
			15056P &            EDWARDIAN PARASOL PINK \\
			10080 &          GROOVY CACTUS INFLATABLE \\
			\bottomrule
		\end{tabular}\label{tab:t10_feat_importance_desc}
	\end{table}
	The details of validating that the linear model assumption for customer value is a reasonable one are provided in section \ref{sec:lasso}. The results of executing these steps are available in \cite{retail_eda_2}. Results showed that a linear model captures the relationship between the customer value and their purchases accurately. A comparison of the error distribution from the model to a normal distribution showed that it is reasonable to assume that the errors from the model estimates follow a normal distribution. 
	\subsection{Matrix Factorization}\label{sec:results_nmf}
\begin{figure}[ht]
		\centering
		\includegraphics[scale=0.33]{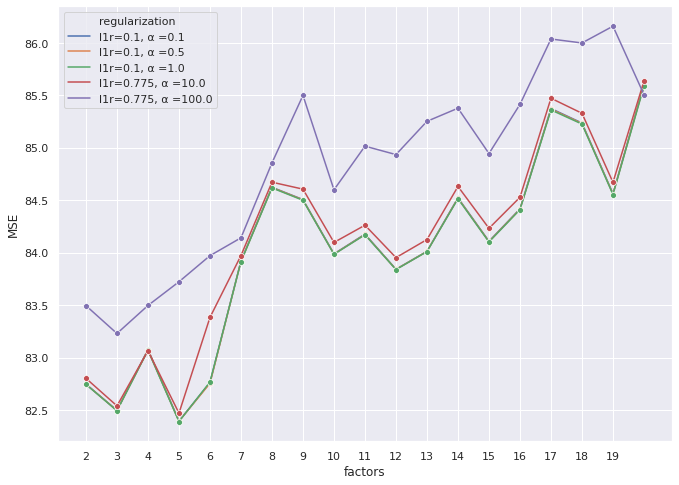}
		\caption{Matrix factorization, choosing the best k.}
		\label{fig:opt_k_mf}	
	\end{figure}

	The flow chart shown in Figure \ref{fig:mf_flowchart} illustrates the steps involved in applying matrix factorization to the purchase data after feature selection. The first step in the process is to determine an \emph{optimal} value of the dimensionality, $k$, the hyperparameters, $\ell^{1}_{\text{ratio}}$ and $\alpha_{m}$ associated with computing the matrix factorization of the matrix $\mathbf{P}{\prime}$. A range of values for these parameters were considered (see Figure \ref{fig:opt_k_mf} for the range of values) and the values that minimized the imputation error on the held out test set was used to develop the matrix factorization model. The results from this experiment are shown in Figure \ref{fig:opt_k_mf}. The best set of parameters from the experiment are shown in Table \ref{tab:mf_opt_params}. The implementation of the experiment is available as a \texttt{colab}\cite{google_colab} notebook\cite{retail_mf}.

	\begin{table}[ht]
		\centering
		\caption{Matrix Factorization Parameters}
		\begin{tabular}{ll}
			\toprule
			NMF parameter &                        Value \\
			\midrule
			$\ell^{1}_{\text{ratio}}$ &     $0.1$ \\
			$\alpha_{m}$ &                      $1.0$ \\
			$k$  &    $5$\\
			\bottomrule
		\end{tabular}\label{tab:mf_opt_params}
	\end{table}

	\subsection{Profiling the Matrix Factorization Solution}\label{sec:results_nmf_profiles}
\begin{figure}[ht]
		\centering
		\includegraphics[scale=0.33]{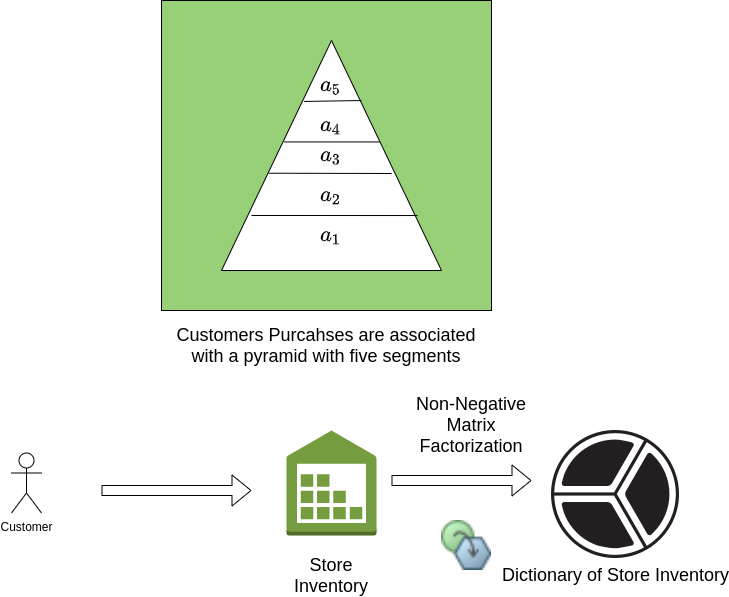}
		\caption{Interpretation of Matrix Factorization.}
		\label{fig:xplainmf}	
	\end{figure}

	Matrix factorization is applied to the incidence matrix with the latent dimension size, $k$, set to $5$ based on the results from the experiment discussed in the previous section. Matrix factorization provides matrices $\mathbf{W}$ and $\mathbf{H}$. The interpretation of these matrices is discussed in section \ref{sec:theoretical_background}. The matrix $\mathbf{H}$ is called the dictionary. The dictionary has $5$ elements. The purchases of the $i^{th}$ customer, $p_i$, can be expressed in terms of his/her affinities towards the dictionary elements, as indicated by Equation \ref{eqn:mf_purchase_approx}. This is illustrated in Figure \ref{fig:xplainmf} where each customer's affinities towards the dictionary elements can be represented by a pyramid.
	
\begin{figure}[ht]
		\centering
		\includegraphics[scale=0.33]{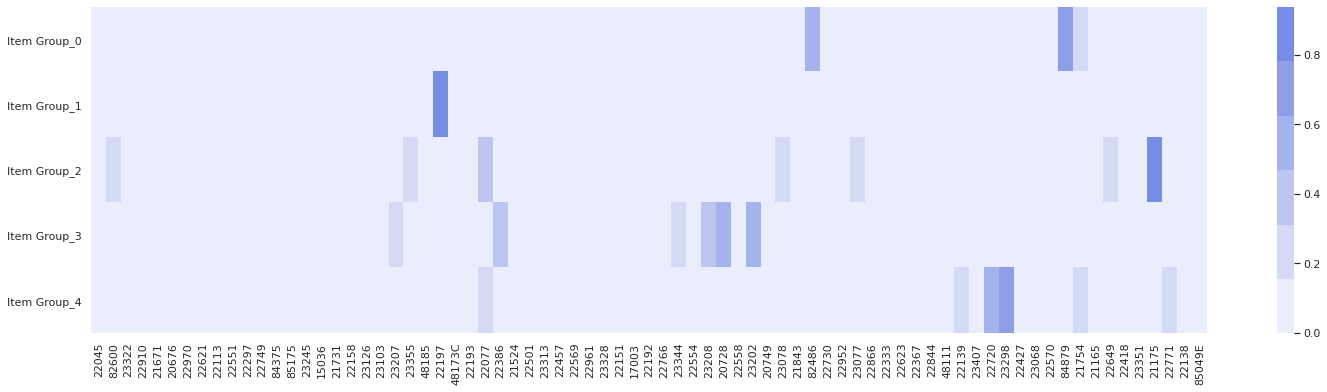}
		\caption{Profile of the frequent shopper dictionary.}
		\label{fig:dictionary_profile}	
	\end{figure}

	 To determine the relative contribution of each inventory item to a dictionary element, we can normalize the rows of the matrix $\mathbf{H}$ to be of unit length. This provides a quick view of the relative importance of the inventory items in a dictionary element. The results of this computation are shown in Figure \ref{fig:dictionary_profile}

	A review of Figure \ref{fig:dictionary_profile} shows that each dictionary element is dominated by a small number of inventory items. Table \ref{tab:dictionary_profile} provides the description of the top 3 inventory for each of the five dictionary elements. The dictionary elements can overlap\footnote{Formulations of non-negative matrix factorization that yield orthogonal basis are also possible}. Note for example that the \emph{christmas bunting} figures both the first and second dictionary elements in Table \ref{tab:dictionary_profile}. The reader may be tempted to compare Table \ref{tab:t10_feat_importance_desc} with Table \ref{tab:dictionary_profile}. It should be noted that dictionary learning is based on the objective function in Equation \ref{eqn:nmfobjective} and is \emph{unsupervised} learning. In contrast, the feature importance shown in Table \ref{tab:t10_feat_importance_desc} is based on \emph{supervised} learning with the objective function described in Equation \ref{eqn:rfmpurchases}. These facts should be kept in mind when interpreting the results. The \emph{customer-value} is not explicitly used in learning the dictionary representation. Exploring this is an area of future work, as discussed in section \ref{sec:related_work}. It is possible to select a specific basis (rather than learn it) when developing the matrix factorization solution. For example, if we like the set of dictionary elements learned in this work, we can apply it in future analysis to develop a solution with a new sample of data, see \cite{manohar2018data} for details. In other words, incorporating prior knowledge about the dictionary is an option with non-negative matrix factorization. To the best of our knowledge, we did not find other work analyzing this data.

	\begin{table}[H]
		\centering
		\caption{Profile of the dictionary of purchases.}
		
		\begin{tabular}{lrl}
			\toprule
			Item\_group &  Item\_ID &                          Description \\
			\midrule
			Item Group\_0 &       11 &    FELTCRAFT PRINCESS CHARLOTTE DOLL \\
			Item Group\_0 &       26 &               6 RIBBONS RUSTIC CHARM \\
			Item Group\_0 &       27 &              JUMBO BAG PINK POLKADOT \\
			Item Group\_1 &       30 &            VINTAGE CHRISTMAS BUNTING \\
			Item Group\_1 &       48 &  3 DRAWER ANTIQUE WHITE WOOD CABINET \\
			Item Group\_1 &       44 &               JUMBO BAG VINTAGE LEAF \\
			Item Group\_2 &       37 &                BLUE DINER WALL CLOCK \\
			Item Group\_2 &       30 &            VINTAGE CHRISTMAS BUNTING \\
			Item Group\_2 &       41 &        LUNCH BAG VINTAGE LEAF DESIGN \\
			Item Group\_3 &       74 &            SCANDINAVIAN REDS RIBBONS \\
			Item Group\_3 &       57 &                DOORMAT 3 SMILEY CATS \\
			Item Group\_3 &       54 &        BOX OF VINTAGE JIGSAW BLOCKS  \\
			Item Group\_4 &        0 &                   SPACEBOY GIFT WRAP \\
			Item Group\_4 &       39 &            JUMBO BAG 50'S CHRISTMAS  \\
			Item Group\_4 &       23 &                       POPCORN HOLDER \\
			\bottomrule
		\end{tabular}\label{tab:dictionary_profile}
	\end{table}
	
\begin{figure}[ht]
		\centering
		\includegraphics[scale=0.33]{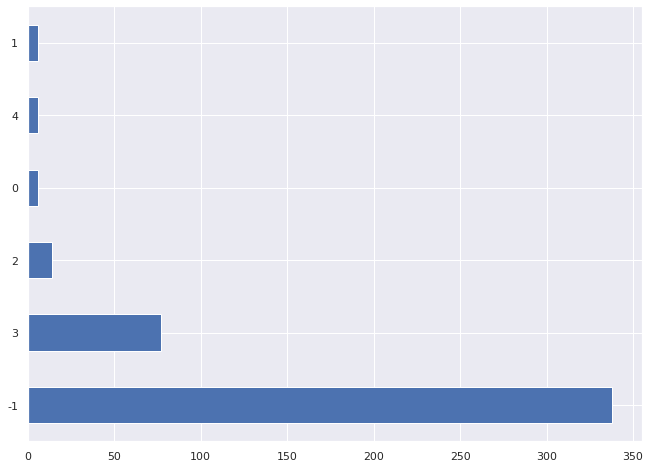}
		\caption{Cluster sizes from \texttt{hdbscan}.}
		\label{fig:cluster_sizes}	
	\end{figure}

	The rows of the matrix, $\mathbf{W}$, represent the affinities of the customers towards the dictionary elements. By clustering the matrix $\mathbf{W}$, we can see how preferences of the store customers towards the dictionary elements group together. The \texttt{hdbscan}\cite{mcinnes2017hdbscan}\cite{campello2013density} algorithm is used for this purpose. \texttt{hdbscan} can discriminate between \emph{noise} and clustering tendency. The algorithm requires us to specify the minimum number of points to consider aggregating into a cluster. Given the size of the frequent shopper group ($447$ customers), this is set to $5$. \texttt{hdbscan} produces $5$ clusters in the data with a large noise cluster ($338$). One of these clusters, Cluster $3$, is large, with $77$ customers. There is a cluster with $14$ elements - Cluster $2$. There are three small clusters, Cluster $0$, Cluster $4$, and Cluster $1$, each with $6$ customers. A size distribution of the clusters is shown in Figure \ref{fig:cluster_sizes}.
	
\begin{figure}[H]
		\centering
		\includegraphics[scale=0.33]{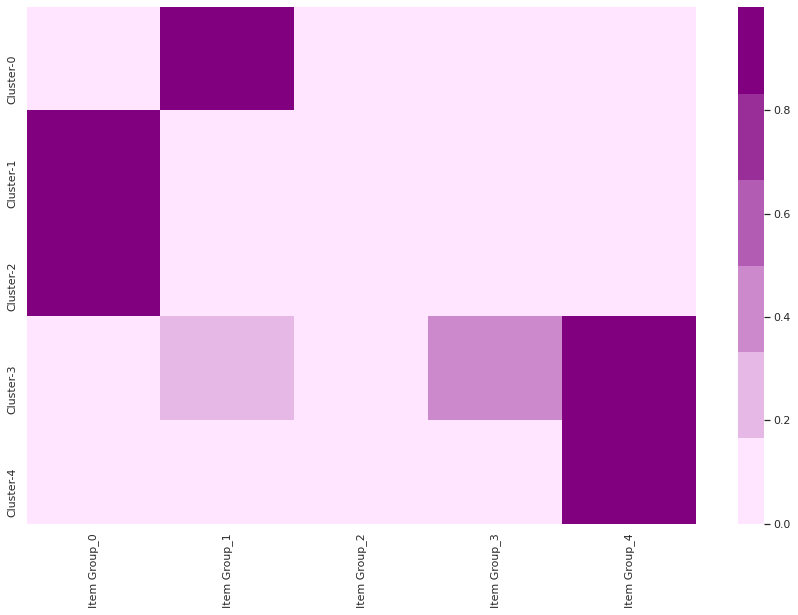}
		\caption{Cluster centroid affinities.}
		\label{fig:cluster_centroid_affinities}	
	\end{figure}

	A review of the results from \texttt{hdbscan} reveals there is a large group of customers ($338$) that we cannot assign a cluster to. This is the noise cluster. The purchase activity of this group does not show clustering tendencies. There is a smaller group of $109$ \footnote{The size of the smaller group with clustering tendencies is simply the difference between the size of the group and the size of the noise cluster.} customers whose purchasing activities show clear clustering tendencies. The cluster centroids of each of these clusters can be computed. Each centroid can then be normalized to unit length. An inspection of the coordinates of the normalized centroids reveals the affinities each cluster has to the dictionary elements. The results from such a computation are shown in Figure \ref{fig:cluster_centroid_affinities}. Cluster $0$ has a clear preference for the dictionary element tagged \texttt{Item Group\_1}. Clusters $1$ and Cluster $2$ have a strong affinity to the dictionary element tagged \texttt{Item Group\_0}. These clusters are over-lapping clusters that are characterized by a strong preference towards dictionary element \texttt{Item Group\_0} and slightly different and small affinities towards the other dictionary elements. Cluster $3$, the large cluster, has affinities for dictionary elements \texttt{Item Group\_1}, \texttt{Item Group\_3} and \texttt{Item Group\_4}. There is a strong preference for \texttt{Item Group\_4} in Cluster $3$. Cluster $4$ is characterized by affinity for dictionary element \texttt{Item Group\_4}. Cluster $4$ does not show strong affinities for \texttt{Item Group\_1} and \texttt{Item Group\_3}. Overall, it is clear that the dictionary element, \texttt{Item Group\_4}, is favored by a majority of customers with clear shopping preferences. The elements of this dictionary element, see Table \ref{tab:dictionary_profile}, capture an important facet of the taste of the frequent shoppers. Ensuring the supply and availability of the inventory items that dominate dictionary element, \texttt{Item Group\_4}, is important to keep store customers satisfied. 
	
	A graph is an excellent data model for this problem. Both the business problem and the results from its analysis have very elegant interpretations as graphs. The customer's purchase activity could be modeled and interpreted using a bipartite graph. Similarly, the results from this analysis also have a natural interpretation as a bipartite graph. The bipartite graph capturing the purchase activity related to the small set of inventory items capturing customer value is one of the results from this analysis. The results from matrix factorization also have an interpretation as a bipartite graph with the customers and the dictionary elements forming the vertex sets. The matrix $\mathbf{W}$ represents the edge weights between the customers and the dictionary vertex sets. These graphs can be stored in a \emph{Graph Database}. For the storage, we used an ArangoDB graph database. The embeddings of the vertex sets can also be stored. This graph representation can be used to capture and query the characteristics of frequent shoppers to the store. For example, an analyst can query this graph to see if a customer is similar to another customer. The data for this work covers a particular period. Similar analyses performed for other periods produce similar results. An analyst can then query these graphs to see understand how customer preferences change over time. For example, an analyst could examine how dictionary profiles and customer cluster profiles change over time (see \cite{retail_mf} for examples of such queries).

	\subsection{ Discussion of Results} \label{sec:discussion_of_results}
	The feature selection models illustrate that a small set of $75$ store inventory items (out of the $2664$ items in the inventory) account for about $86\%$ of the variation in customer value. Since a linear model was used, the elements of this set can be ordered based on their associated coefficients in the model. The first $50$ of these elements account for most of the customer value explained by the linear model. Managing the inventory and restocking of these elements is critical from the standpoint of store operations. Matrix factorization groups these $75$ inventory items into dictionary elements. Analysis of the grouping showed that each dictionary element is characterized by a small set of inventory items. The dictionary elements are not orthogonal. This implies that there can overlap between the items that characterize the dictionary elements. Formulations of non-negative matrix factorization that yield orthogonal bases are possible, for example, \cite{asteris2015orthogonal}. A customer's purchase can be viewed in terms of affinity towards these dictionary elements (Equation \ref{eqn:mf_purchase_approx}). Clustering of customer affinities revealed that there is a rather large group whose tastes do not align with other customers. There is a small group of $109$ customers whose taste aligns with other customers. A profile of the cluster centroids provided the preferences of each of these groups to the dictionary elements. The characterization of frequent shoppers developed in this work can be stored and analyzed as a graph in \emph{Graph database} such as \texttt{ArangoDB}. The \emph{embeddings} of the vertices of the graph are also stored as properties of graph nodes. These embeddings can be used to develop applications such as recommendation systems \cite{messica_session_based}. In summary, the methodology applied in this work provides embeddings, insights, and interpretable models! 
	
	\section{Conclusion} \label{sec:conclusion}
	This work uses sparsity-seeking models to develop the characterization of the frequent shoppers to a real-world online store. The models used were interpretable and yielded insights about frequent customers to the store. The theoretical properties of each of the methods applied are well established. Methods to apply these techniques to large datasets are also established. The perspective in this work is analysis as opposed to forecasting or prediction. To the best of our knowledge, there is no other work analyzing the group of frequent shoppers associated with this data from a shopping preferences perspective. The use of interpretable models permits us to identify a set of facts that form a baseline for future work on this data. Exploring supervised matrix factorization is an area of future work. A comparison of the results from this work to results from graph reduction approaches, such as coarsening, would be interesting. A comparison of the analytical techniques from the perspective of information provided, cost and computational characteristics etc. is an area for future work. Characterization of the wholesale shoppers and infrequent shoppers is also an area of future work. 

	\bibliographystyle{unsrt}           
	\bibliography{references}

\begin{thebibliography}{10}

\bibitem{hbr_cust_pref}
{Davenport, Thomas and Dallamule Leonardo and Lucker John}.
\newblock {Know What Your Customers Want Before They Do}, 2011.

\bibitem{retail_wire_survey}
{Phibbs, Bob}.
\newblock {What do shoppers really want? Do retailers have a clue?}, 2021.

\bibitem{hastie2019statistical}
Trevor Hastie, Robert Tibshirani, and Martin Wainwright.
\newblock {\em Statistical learning with sparsity: the lasso and
  generalizations}.
\newblock Chapman and Hall/CRC, 2019.

\bibitem{shmueli2010explain}
Galit Shmueli.
\newblock To explain or to predict?
\newblock {\em Statistical science}, 25(3):289--310, 2010.

\bibitem{moitra2018algorithmic}
Ankur Moitra.
\newblock {\em Algorithmic aspects of machine learning}.
\newblock Cambridge University Press, 2018.

\bibitem{dua_2019}
Dheeru Dua and Casey Graff.
\newblock {UCI} machine learning repository, 2017.

\bibitem{nlp_ir_online_ref}
{Manning, Christopher and Raghavan, Prabhakar and Schütze, Hinrich}.
\newblock {An example information retrieval problem}, 2008.

\bibitem{DBLP:books/daglib/0021593}
Christopher~D. Manning, Prabhakar Raghavan, and Hinrich Sch{\"{u}}tze.
\newblock {\em Introduction to information retrieval}.
\newblock Cambridge University Press, 2008.

\bibitem{eggert2004sparse}
Julian Eggert and Edgar Korner.
\newblock Sparse coding and nmf.
\newblock In {\em 2004 IEEE International Joint Conference on Neural Networks
  (IEEE Cat. No. 04CH37541)}, volume~4, pages 2529--2533. IEEE, 2004.

\bibitem{lecturenotes_nicolai}
{Nicolai Meinshausen}.
\newblock {Multivariate Statistics – non-negative matrix factorisation and
  sparse dictionary learning}, 2021.

\bibitem{google_colab}
{Google}.
\newblock {Google Colab}, 2021.

\bibitem{retail_eda_1}
{ArangoDB}.
\newblock {Exploratory Data Analysis of a Retail Dataset}, 2021.

\bibitem{fox2019regression}
John Fox.
\newblock {\em Regression diagnostics: An introduction}.
\newblock Sage publications, 2019.

\bibitem{box1964analysis}
George~EP Box and David~R Cox.
\newblock An analysis of transformations.
\newblock {\em Journal of the Royal Statistical Society: Series B
  (Methodological)}, 26(2):211--243, 1964.

\bibitem{chen2012data}
Daqing Chen, Sai~Laing Sain, and Kun Guo.
\newblock Data mining for the online retail industry: A case study of rfm
  model-based customer segmentation using data mining.
\newblock {\em Journal of Database Marketing \& Customer Strategy Management},
  19(3):197--208, 2012.

\bibitem{lloyd1982least}
Stuart Lloyd.
\newblock Least squares quantization in pcm.
\newblock {\em IEEE transactions on information theory}, 28(2):129--137, 1982.

\bibitem{breiman2017classification}
Leo Breiman, Jerome~H Friedman, Richard~A Olshen, and Charles~J Stone.
\newblock {\em Classification and regression trees}.
\newblock Routledge, 2017.

\bibitem{mcinnes2017accelerated}
Leland McInnes and John Healy.
\newblock Accelerated hierarchical density based clustering.
\newblock In {\em 2017 IEEE International Conference on Data Mining Workshops
  (ICDMW)}, pages 33--42. IEEE, 2017.

\bibitem{austin2018fully}
Woody Austin, Dylan Anderson, and Joydeep Ghosh.
\newblock Fully supervised non-negative matrix factorization for feature
  extraction.
\newblock In {\em IGARSS 2018-2018 IEEE International Geoscience and Remote
  Sensing Symposium}, pages 5772--5775. IEEE, 2018.

\bibitem{choo2015weakly}
Jaegul Choo, Changhyun Lee, Chandan~K Reddy, and Haesun Park.
\newblock Weakly supervised nonnegative matrix factorization for user-driven
  clustering.
\newblock {\em Data mining and knowledge discovery}, 29(6):1598--1621, 2015.

\bibitem{blei2003latent}
David~M Blei, Andrew~Y Ng, and Michael~I Jordan.
\newblock Latent dirichlet allocation.
\newblock {\em the Journal of machine Learning research}, 3:993--1022, 2003.

\bibitem{andreas_graph_coarsening}
Andreas Loukas.
\newblock Graph reduction with spectral and cut guarantees.
\newblock {\em Journal of Machine Learning Research}, 20(116):1--42, 2019.

\bibitem{scikit-learn}
F.~Pedregosa, G.~Varoquaux, A.~Gramfort, V.~Michel, B.~Thirion, O.~Grisel,
  M.~Blondel, P.~Prettenhofer, R.~Weiss, V.~Dubourg, J.~Vanderplas, A.~Passos,
  D.~Cournapeau, M.~Brucher, M.~Perrot, and E.~Duchesnay.
\newblock Scikit-learn: Machine learning in {P}ython.
\newblock {\em Journal of Machine Learning Research}, 12:2825--2830, 2011.

\bibitem{tibshirani1996regression}
Robert Tibshirani.
\newblock Regression shrinkage and selection via the lasso.
\newblock {\em Journal of the Royal Statistical Society: Series B
  (Methodological)}, 58(1):267--288, 1996.

\bibitem{tibshirani2013lasso}
Ryan~J Tibshirani.
\newblock The lasso problem and uniqueness.
\newblock {\em Electronic Journal of statistics}, 7:1456--1490, 2013.

\bibitem{chen1998application}
Scott~Shaobing Chen and David~L Donoho.
\newblock Application of basis pursuit in spectrum estimation.
\newblock In {\em Proceedings of the 1998 IEEE International Conference on
  Acoustics, Speech and Signal Processing, ICASSP'98 (Cat. No. 98CH36181)},
  volume~3, pages 1865--1868. IEEE, 1998.

\bibitem{kfoldcrossval}
{Tibshirani, Ryan}.
\newblock {K-Fold Cross Validation}, 2021.

\bibitem{retail_eda_2}
{ArangoDB}.
\newblock {Feature Selection Using the LASSO on a Retail Dataset}, 2021.

\bibitem{johnson2019billion}
Jeff Johnson, Matthijs Douze, and Herv{\'e} J{\'e}gou.
\newblock Billion-scale similarity search with gpus.
\newblock {\em IEEE Transactions on Big Data}, 2019.

\bibitem{hamon2016convex}
Ronan Hamon, Valentin Emiya, and C{\'e}dric F{\'e}votte.
\newblock Convex nonnegative matrix factorization with missing data.
\newblock In {\em 2016 IEEE 26th International Workshop on Machine Learning for
  Signal Processing (MLSP)}, pages 1--6. IEEE, 2016.

\bibitem{lee1999learning}
Daniel~D Lee and H~Sebastian Seung.
\newblock Learning the parts of objects by non-negative matrix factorization.
\newblock {\em Nature}, 401(6755):788--791, 1999.

\bibitem{essid2014tutorial}
Slim Essid and Alexey Ozerov.
\newblock A tutorial on nonnegative matrix factorisation with applications to
  audiovisual content analysis.
\newblock In {\em Tutorial at ICME 2014}, 2014.

\bibitem{DBLP:journals/pami/DingLJ10}
Chris H.~Q. Ding, Tao Li, and Michael~I. Jordan.
\newblock Convex and semi-nonnegative matrix factorizations.
\newblock {\em {IEEE} Trans. Pattern Anal. Mach. Intell.}, 32(1):45--55, 2010.

\bibitem{fevotte2011algorithms}
C{\'e}dric F{\'e}votte and J{\'e}r{\^o}me Idier.
\newblock Algorithms for nonnegative matrix factorization with the
  $\beta$-divergence.
\newblock {\em Neural computation}, 23(9):2421--2456, 2011.

\bibitem{DBLP:journals/datamine/Domingos99}
Pedro~M. Domingos.
\newblock The role of occam's razor in knowledge discovery.
\newblock {\em Data Min. Knowl. Discov.}, 3(4):409--425, 1999.

\bibitem{DBLP:journals/cacm/Domingos12}
Pedro~M. Domingos.
\newblock A few useful things to know about machine learning.
\newblock {\em Commun. {ACM}}, 55(10):78--87, 2012.

\bibitem{retail_mf}
{ArangoDB}.
\newblock {Matrix Factorization of Frequent Shopper Purchases}, 2021.

\bibitem{manohar2018data}
Krithika Manohar, Bingni~W Brunton, J~Nathan Kutz, and Steven~L Brunton.
\newblock Data-driven sparse sensor placement for reconstruction: Demonstrating
  the benefits of exploiting known patterns.
\newblock {\em IEEE Control Systems Magazine}, 38(3):63--86, 2018.

\bibitem{mcinnes2017hdbscan}
Leland McInnes, John Healy, and Steve Astels.
\newblock hdbscan: Hierarchical density based clustering.
\newblock {\em Journal of Open Source Software}, 2(11):205, 2017.

\bibitem{campello2013density}
Ricardo~JGB Campello, Davoud Moulavi, and J{\"o}rg Sander.
\newblock Density-based clustering based on hierarchical density estimates.
\newblock In {\em Pacific-Asia conference on knowledge discovery and data
  mining}, pages 160--172. Springer, 2013.

\bibitem{asteris2015orthogonal}
Megasthenis Asteris, Dimitris Papailiopoulos, and Alexandros~G Dimakis.
\newblock Orthogonal nmf through subspace exploration.
\newblock {\em Advances in Neural Information Processing Systems}, 28:343--351,
  2015.

\bibitem{messica_session_based}
Asnat Greenstein-Messica, Lior Rokach, and Michael Friedman.
\newblock Session-based recommendations using item embedding.
\newblock In {\em Proceedings of the 22nd International Conference on
  Intelligent User Interfaces}, IUI '17, page 629–633, New York, NY, USA,
  2017. Association for Computing Machinery.

\end{thebibliography}

\end{document}